# The functional and temporal roles of gaze evolve across the phases and constraints of multi-stage robot-mediated manipulation

Gaze Reorganization during Robot-Mediated Manipulation

The human visuomotor system dynamically reconfigures across task phases, flexibly alternating between exploratory, predictive, and online monitoring functions.

Manuela Uliano
The BioRobotics Institute, Scuola Superiore Sant'Anna; Pisa, Italy
Department of Excellence in Robotics and AI, Scuola Superiore Sant'Anna; Pisa, Italy
manuela.uliano@santannapisa.it

Silvia Fattorini
The BioRobotics Institute, Scuola Superiore Sant'Anna; Pisa, Italy
Department of Excellence in Robotics and AI, Scuola Superiore Sant'Anna; Pisa, Italy
silvia.fattorini@santannapisa.it

Marco Controzzi*
The BioRobotics Institute, Scuola Superiore Sant'Anna; Pisa, Italy
Department of Excellence in Robotics and AI, Scuola Superiore Sant'Anna; Pisa, Italy
marco.controzzi@santannapisa.it

* corresponding author

Goal-directed eye movements are a fundamental component of visuomotor control, enabling humans to anticipate and guide their actions. For this reason, they are increasingly used in human-robot interaction to estimate users' goals. However, during manipulation, fixations may reflect either an intended future action or the need to visually monitor the robotic proxy due to altered embodiment. How predictive and monitoring-related gaze are organized across the different phases of a constrained robot-mediated manipulation remains unclear.
Here we address this question by investigating gaze behavior during goal-directed telemanipulation to characterize how visuomotor control adapts to altered embodiment in a multi-stage task. Our findings show that gaze remains strongly aligned with task goals, preserving its predictive role even during robot-mediated manipulation. At the same time, gaze frequently alternates between the robotic end-effector and the manipulated object, revealing increasing online monitoring. The presence and geometry of obstacles modulate the timing and distribution of these fixations, delaying attention to the final target when intermediate constraints become more demanding. These findings show that predictive gaze is not lost under altered embodiment but reorganized in response to changes in sensory feedback and control demands. More broadly, they highlight the flexibility of the human visuomotor system when the natural sensorimotor coupling is disrupted and suggest that gaze should be interpreted contextually rather than treating every fixation as direct evidence of user intention in human-robot interaction.

## 1 INTRODUCTION

Goal-directed actions rely on the coordinated control of eye and hand movements. During natural object manipulation, gaze typically precedes the hand, providing predictive information that guides contact points and supports efficient motor control. Indeed, eye movements anticipate manipulative actions by approximately 500 ms [1-3] and prioritize task-relevant objects over their visual salience [1, 2, 4]. Rather than continuously monitoring the hand or the moving object, humans generally rely on proprioception and peripheral vision while directing gaze toward future contact points, particularly the anticipated location of the index finger [5, 6]. Gaze behavior is also shaped by object familiarity, with unfamiliar manipulanda eliciting more visual inspection [7], as well as by task complexity and cognitive load, which lead to more erratic and less stable gaze patterns [8].

Most evidence on this predictive coordination comes from natural manipulation, where perception and action are tightly coupled through an intact sensorimotor loop. However, interactions are not always direct. In many human-computer interaction settings, users act through a virtual intermediate representation such as a cursor or a virtual hand. These systems alter the relationship between biological motion, visual feedback, proprioceptive information, and task outcomes, potentially changing how visual information supports motor control. Research on this topic suggests that users of these systems rely more on visual monitoring when the mapping between their movement and the controlled proxy is unfamiliar or uncertain [9, 10]. Studies in virtual reality have demonstrated that the mediation does not eliminate the predictive role of gaze, while also emphasizing its importance for online monitoring and control during object manipulation [11-13]. More recently, direct comparisons between controller-mediated virtual manipulation and direct manipulation have shown that anticipatory eye-hand coordination is largely preserved, whereas users spend more time visually monitoring manipulated objects, likely compensating for reduced haptic information [14].

Among intermediate representations, robot teleoperation represents one of the most demanding, as in this case users control a remote but physical effector while coping with latency, altered kinematics, limited haptic information and mismatches between human and robotic degrees of freedom, leading users to rely primarily on visual feedback [15-20]. Similar challenges arise in prosthesis use [21-23], where sensory feedback and mapping between biological and controlled movements are also altered. Unlike teleoperated robots, however, the prosthetic effector is physically attached to the user's body and moves within the user's peripersonal space, leading to important differences in sensorimotor integration. Across these settings, predictive and monitoring-related gaze have both been observed, but their phase-dependent organization during constrained robot-mediated manipulation is not yet well understood. Several studies have reported increased attention toward the controlled effector and the manipulated object, consistent with a greater dependence on visual feedback during action execution [24-27]. Conversely, other studies have suggested that anticipatory fixations toward future action goals remain present, particularly as users become familiar with the control mapping [3, 28]. However, this finding is not consistent across teleoperation studies, as some of them have reported no reliable anticipatory gaze toward task goals [27, 29].

Despite extensive work on gaze during mediated manipulation, differences in terms of tasks, control mappings, and experimental protocols make it difficult to determine how robotic mediation reshapes visuomotor coordination. More specifically, it remains unclear how predictive and monitoring-related fixations coexist across the sequential phases of a constrained manipulation task. A clearer understanding of how gaze supports robot-mediated manipulation may inform shared and assistive control frameworks, which combine user commands with autonomous robot behavior [30-33]. In these contexts, gaze can provide implicit evidence about user intention [32], but fixations on the robotic hand, manipulated object, or an intermediate obstacle may instead reflect online monitoring or constraint management. Its interpretation therefore depends on both the functional role of the fixated element and the current phase of the task [33]. Here, we address this gap by adapting a well-established paradigm of natural manipulation [4] to a teleoperated setting. Participants remotely controlled a robotic arm and hand to perform a goal-directed manipulation while viewing the remote scene through a virtual reality headset. We varied obstacle presence and geometry across three conditions to manipulate intermediate control constraints and assess how gaze was distributed among the final goal, the obstacle, the manipulated object, and the robotic hand (Figure 1). We asked whether predictive gaze remains present during robot-mediated manipulation, whether it coexists with increased proxy monitoring, and whether the balance between these functions changes across task phases and obstacle conditions. This design provides a structured reference to previously characterized natural eye-hand coordination while enabling a phase-resolved analysis of gaze during teleoperation.

Building on the natural manipulation benchmark provided by Johansson et al. [4], we characterize how predictive and monitoring-related gaze are jointly organized across the different phases of robot-mediated manipulation. Results show that predictive gaze is preserved during robot-mediated manipulation while reorganizing its phase-dependent allocation toward increased monitoring of the end-effector and manipulated object. More broadly, these findings clarify how the human visuomotor system adapts when natural sensory motor coupling is disrupted and establish gaze as a robust signal of action intention in human-robot interaction.

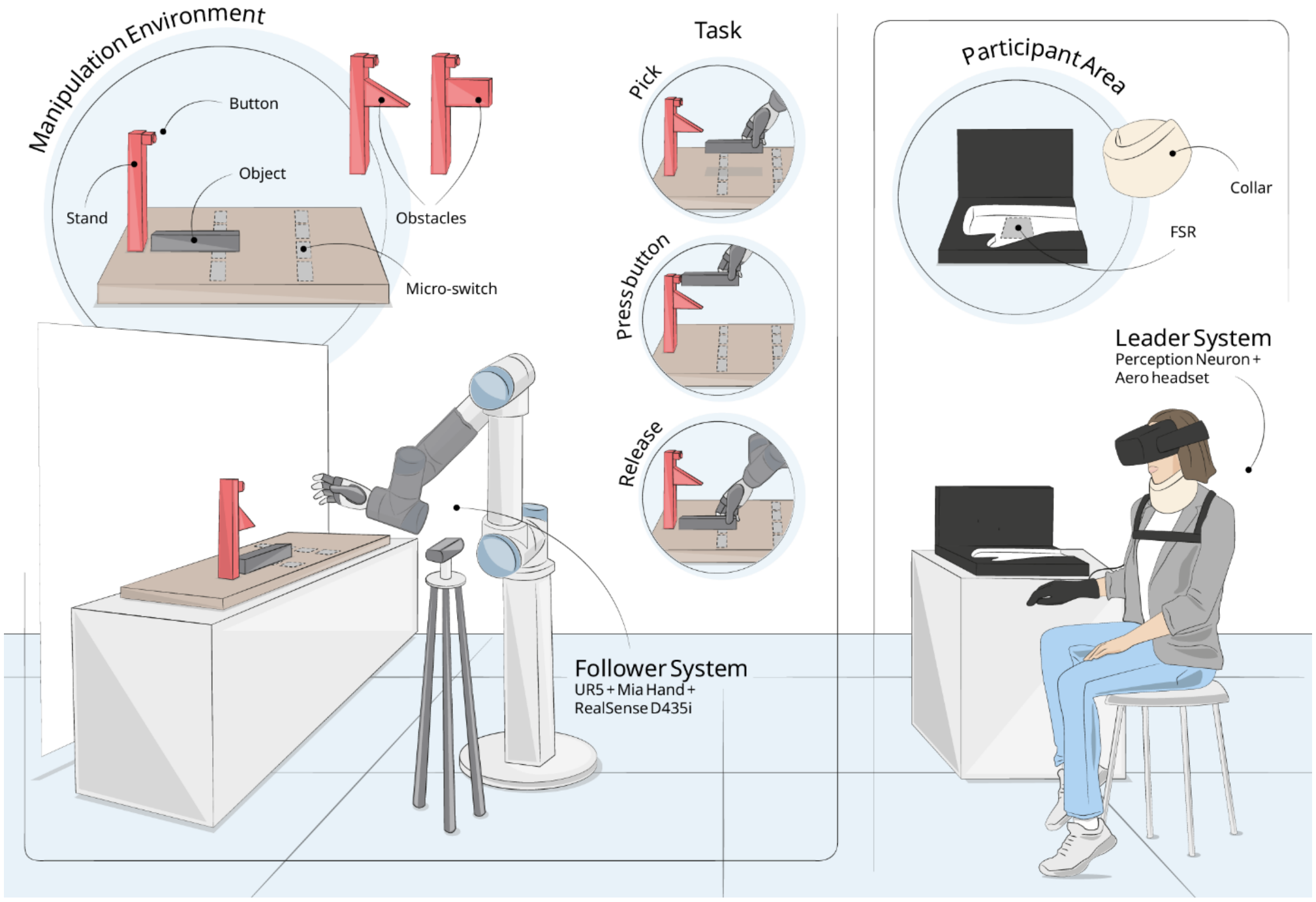

**Fig. 1. Experimental setup and task.** The teleoperation platform comprised a leader and a follower system. The leader system included a wearable motion capture device and a virtual reality headset; the follower system consisted of a robotic arm equipped with an anthropomorphic robotic hand as end-effector, together with an RGB-D camera to capture the remote manipulation environment. To ensure a visually homogeneous appearance and avoid eye-catching features, the robotic arm was covered with a tight-fitting grey fabric. The manipulation environment included a micro-switch equipped support, an object, a stand, a button, and two types of obstacles with distinct shapes: rectangular and right-angled triangular. The participant area consisted of a hand support and a cervical collar. The hand support was placed on a table to the participant's right and embedded a contact Force Sensitive Resistor (FSR) sensor to detect hand contact and release. Participants teleoperated the robot to pick the object from the support using the index and thumb fingers and to press

the button with the object's left end. After completing this action, they released the object onto the support and repositioned their right hand on the hand support.

## 2 MATERIALS AND METHODS

### 2.1 Participants

Eleven participants (right-handed, 6 males and 5 females; mean age, 27.5; SD, 1.9) were enrolled in the experiment. All participants were healthy, reported normal vision, were naïve teleoperators and were not aware of the purpose of the experiment. All participants took part on a voluntary basis and gave their signed consent to the participation. Informed consent in accordance with the Declaration of Helsinki was obtained from each participant before conducting the experiment. This study was approved by the local ethical committee of the Scuola Superiore Sant'Anna, Pisa, Italy (approval n. 21/2022).

### 2.2 Experimental design

#### *2.2.1 Experimental setup*

The experimental setup consisted of the teleoperation platform, the manipulation environment, and the participant area (Figure 1).

The teleoperation platform comprised the leader and the follower systems. The leader system involved a wearable motion capture device (Perception Neuron 32 V2, Noitom) and a virtual reality headset (Aero, Varjo). The Perception Neuron suit tracked the participant's arm and hand movement at a frequency of 120 Hz, enabling direct control of the robotic arm and hand. Meanwhile, the Aero headset projected the remote environment to the participant and tracked the participant's gaze at a frequency of 200 Hz. On a Windows 11 workstation, the Axis Neuron software was used to process motion capture data, while the Varjo Base software and the Unity editor managed the Aero headset and its virtual reality application. The follower system consisted of a 6-axis industrial robotic arm (UR5, Universal Robots) equipped with an anthropomorphic robotic hand (Mia Hand, Prensilia S.r.l.) as its end-effector. To ensure visual homogeneity and avoid eye-catching features, a black robotic hand was used, and the robotic arm was covered with a tight-fitting grey fabric. An RGB-D camera (D435i, Intel RealSense) was mounted on a tripod 50 cm from the workplane, providing a two-dimensional (2D) frontal view and approximating the typical human eye-hand distance during object manipulation. The camera captured and streamed the remote environment to the participant at 60 fps and a resolution of 960x540 pixels. At the beginning of each trial, the remote environment was not visible to the participant, and a black screen with a fixation dot was presented at coordinates (580, 220) pixels in the image coordinate system, with the origin at the upper-left corner and axes extending rightward and downward. The follower system components were managed through the Robot Operating System (ROS) framework on a computer running Ubuntu 20.04. The video captured by the RGB-D camera was streamed to the Aero headset, from ROS to Unity, resulting in a latency of $62 \pm 6$ ms. Simultaneously, gaze data from the Aero headset were transmitted from Unity to ROS for data recording.

The manipulation environment reproduced the setup realized by Johansson et al. [4], applying a twofold scaling factor to prevent the task from becoming overly precise and demanding under teleoperation. Despite this adjustment, proportions between lengths and distances were preserved, as well as the colors of all elements. The manipulation environment consisted of a sensor-equipped support, a manipulation bar, a stand, a target area, and two types of obstacles with distinct shapes, rectangular and right-angled triangular (Figure 1). The sensor-equipped support (45x20 cm) was made of pre-cut foam containing eight embedded micro-switches (SS-01GL13-E, Omron) to detect the contact between the manipulation bar and the support. These sensors were arranged in two vertical rows (from a top-view perspective), each containing four sensors. The rows were spaced 13 cm apart, with a 1.5 cm gap between adjacent sensors within the same row. The spacing ensured that at least one sensor was fully activated regardless of the bar's position on the support. A beige drape covered the pre-cut foam to prevent these elements from becoming visually distracting. The grey manipulation bar (160x40x40 mm) was three-dimensionally (3D) printed, with its mass (40 g) specifically chosen to reliably activate the micro-switch sensors embedded in the sensor-equipped support while preventing unwanted object flipping if grasped at its extremities. The red wooden stand (260x30x40 mm) was placed on the left side of the support and fixed to an aluminum breadboard through a designed housing. The target area, anchored to the stand and positioned 236 mm from the sensor-equipped support, consisted of a housing (24x24x24 mm) containing a red button. For the experiment, two obstacles with different geometries were employed: a rectangular obstacle (100x80x40 mm) and a right-angled triangular obstacle (100x80x40 mm). Both obstacles were placed 120 mm from the sensor-equipped support and adhered to the stand thanks to two small magnets, allowing easy replacement. The target area and the obstacles were 3D printed using red filament. A solid-colored drape was held by two tripods in the background to prevent external elements from distracting the participant and attracting gaze.

The participant area consisted of a hand support and a cervical collar. The hand support was placed on a table to the participant's right and embedded a contact Force Sensitive Resistor (FSR) sensor (FSR01CE, Ohmite) to detect hand contact and release. It was made from a polystyrene block shaped to fit the hand, ensuring consistent placement and reliable sensor activation even without visual guidance. To further assist hand positioning, a vertically oriented polystyrene block was placed next to the support, providing a tactile reference that guided participants to the correct resting position. Since the protocol required participants to keep their heads still, the RGB-D camera was kept stationary and participants wore a cervical collar (Cy C3, Morsa Cyberg) to help maintain head stability.

All sensors in the experimental setup were interfaced to the robot control box and managed through ROS. The setup components remained fixed throughout the experiment, ensuring identical conditions across all participants.

#### *2.2.2 Experimental protocol*

The experimental protocol closely replicated the methodology outlined by Johansson et al. [4]. We retained the core sequence of grasping a bar, moving it to a target, and returning it to its initial support in order to facilitate comparison with gaze behavior previously reported during natural object manipulation. In the present study, however, the task was performed through a robot-mediated teleoperation system.

Before data collection, after wearing the Perception Neuron suit and the Aero headset, the participant performed the required calibrations and familiarization with the teleoperation platform. At the beginning of the experiment, the participant was seated with their right hand resting on the hand support, fixating a red dot displayed on a black screen in the Aero headset, while the robotic hand remained in its home pose outside the RGB-D camera's field of view. Each trial began with an auditory cue, after which the black screen disappeared to reveal the manipulation environment through the headset. Upon cue onset, the participant was asked to teleoperate the robot at their own pace to grasp the manipulation bar from the support with the index and thumb fingers and press the button in the target area with the left end of the bar. An additional auditory cue indicated the contact with the target area, which occurred when the button was fully pressed with the manipulation bar. Once this action was completed, the participant replaced the bar on the support and repositioned their right hand on the hand support. At this point, the task was completed, the black screen with the fixation dot was displayed again to the participant, and the robotic system autonomously reset to its home pose. Since the scene displayed in the headset was 2D and depth cannot be perceived, the translational movement of the robotic hand was constrained to a vertical plane intersecting the minor axis of the object (i.e., workplane), while rotations were unrestricted. Throughout the trial, white noise attenuated environmental sounds while preserving the audibility of the task-related auditory cues.
Participants performed 4 repetitions of the task in each of the 3 experimental conditions, which differed based on the presence and shape of an obstacle: no-obstacle condition (NONE), rectangular-obstacle condition (RECT), and triangular-obstacle condition (TRI). The task was first performed under the NONE condition, followed by a randomized order of RECT and TRI conditions. Across trials, the distance between the bar's left tip on the support and the stand was randomly varied between 1.6 and 9.2 cm. A 5 to 10 min break was scheduled between conditions. In total, including instructions and consent procedures, the experiment required approximately 60 minutes.

*2.2.3 Landmark zones*

To preserve comparability with the natural-manipulation paradigm of Johansson et al. [4], we deliberately adopted the same general landmark framework while adapting the operational definitions to the present teleoperation setup and introducing a robot-specific hand landmark. Landmark zones were specified for the grasp site, the left tip of the bar, the target, the task-relevant region of each obstacle, and the support surface. Each landmark was surrounded by a region extending 4 cm in every direction. The grasp landmark was defined by the position of the robotic index finger, expressed in object coordinates, at the instant when the bar started to move. This event was detected when all micro-switches embedded in the support were released. The bar's left tip landmark was defined as the midpoint of the bar's left end, whereas the target landmark was placed at the center of the target area's right surface. Obstacle landmarks were defined in an analogous manner: for the triangular obstacle, the relevant landmark was defined at the projecting tip; for the rectangular obstacle, it was represented by a vertical line aligned with the obstacle's right edge. The support surface landmark was defined as a horizontal line at the median vertical coordinate of the bar when resting on the table. Finally, we introduced the hand landmark delimited by the edges of the robotic hand.

*2.2.4 Phases of the trial*

We retained an event-based segmentation of the manipulation sequence to enable comparison with the analysis introduced by Johansson et al. [4]. The original framework was expanded and adapted to account for the teleoperation interface, the appearance and disappearance of the robotic hand from the camera view, and the events recorded by the sensors in the present setup. Each trial was therefore divided into 12 consecutive phases (Figure 2). (1) The *Pre-entry* began when the black screen disappeared and ended when the onset of hand movement was detected by FSR sensor; (2) *entry* extended from the end of the pre-entry phase to the first appearance of the robotic hand's index finger in the RGB-D camera image; (3) *reach* started when the index finger became visible and continued until it came within 5 cm of the point at which its trajectory initially entered into the outline of the bar; (4) *adjustment* covered the remaining approach to the grasp site and ended when the distance between the robotic index finger and the grasp site fall below 5 cm; (5) *grasp* began at the end of the adjustment phase and ended when the bar started to move, identified by the release of all micro-switches embedded within the support; (6) *up* extended from the end of the grasp phase until the distance between the bar's left tip and the target became smaller than 3 cm; (7) *to target* comprised the final approach to and the interaction with the target, ending when the button was released; (8) *from target* began at the button release and continued until the distance between the bar's left tip and the target exceeds 3 cm; (9) *down* covered the return movement until the distance between the bar's centroid and the centroid position recorded while the bar rested on the support fell below 3 cm; (10) *replace* extended from the end of the down phase until the bar was placed on the sensor-equipped support, as signaled by the activation of at least one micro-switch; (11) *exit* began after the bar replacement and ended when the robotic index finger was no longer visible to the RGB-D camera image; (12) *reset* covered the final interval until the participant repositioned the right hand on the hand support, as detected by FSR activation.

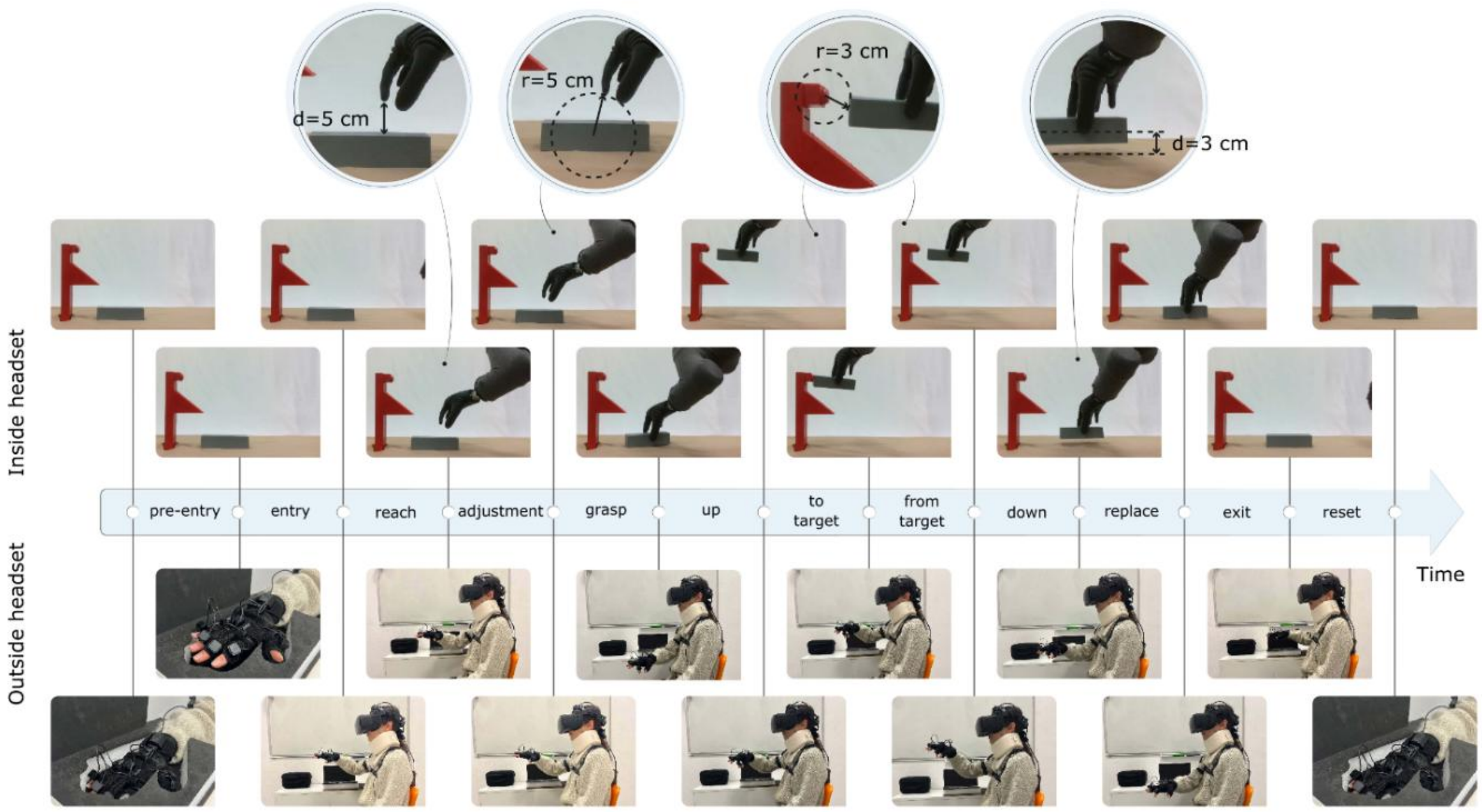

**Fig. 2. Phases of the trial.** The trial is divided into 12 consecutive phases. The figure illustrates the onset and the end of each phase, with the corresponding phase name reported along the timeline. For each phase, the visual content presented inside the virtual reality headset, as well as the events occurring outside the headset, are shown at both the beginning and the end of the phase.

## 2.3 Data analysis

### *2.3.1 Data processing*

To determine which landmark was attended during each fixation, gaze coordinates and landmark positions were expressed in 2D pixel coordinates on the camera image plane. Gaze movements were first classified into fixations, saccades, smooth pursuits, or noise using a deep learning-based algorithm [35]. Subsequently, consecutive samples labelled as fixations were averaged to obtain a single gaze value per fixation, which was then projected onto the camera image plane to obtain 2D pixel coordinates. To account for gaze uncertainty, each fixation was represented as a circular region, termed gaze dot, whose diameter was defined by the Aero headset's tracking accuracy. Static landmarks (i.e., obstacles, target, and support surface) were defined as a priori in pixel coordinates, whereas dynamic landmarks required additional processing. The bar's left tip was identified in each frame in 2D pixel coordinates on the camera image plane using the DeepLabCut software package [36]. Robot-related landmarks required estimating their pixel locations from robot kinematics. A ChArUco board was used to estimate camera intrinsics and the camera pose relative to the board reference frame [37], while a custom probe mounted on the robot wrist was used to estimate the transformation between the robot and board reference frames. The resulting robot to camera transformation was then used to project robot-related 3D positions onto the camera image plane. The hand landmark was defined as an approximate 2D bounding box enclosing the robotic hand, constructed from the known position of the robot's wrist and refined via color-based segmentation exploiting the contrast between the black robotic hand and the elements of the experimental setup. Finally, the grasp site landmark was defined as the 2D pixel coordinates of the robot's index fingertip at grasp onset (i.e., start of grasp phase). Because the index finger behaves as a rigid body, the fingertip position was defined as a fixed offset from a reference frame whose origin was on the robot's index finger metacarpal.

Fixations were automatically assigned to landmarks using an overlap criterion: a fixation was associated with a landmark when the gaze dot overlapped its area; when the fixation gaze dot overlapped multiple landmark areas, the fixation was assigned to all of them. Because color-based segmentation was unreliable during grasping and object transport due to similar hand and bar colors, landmark assignments were visually inspected by the experimenter to ensure results reliability. Fixations associated with multiple landmarks were analyzed differently depending on the specific analysis. In Section 3.2, only fixations uniquely assigned to a single moving landmark (i.e., the grasp site, bar tip, or hand) were considered. In Section 3.3, gaze shifts were defined as transitions between consecutive fixation landmarks; when a fixation was associated with multiple landmarks, all possible landmark sequences preserving fixation order were considered. In all other analyses, fixations associated with multiple landmarks equally contributed to all corresponding landmarks.

Trial phases were automatically identified using an algorithm that combined discrete event timings from the setup sensors with the 2D pixel coordinates of setup elements (Section 2.2.4).

All data were recorded in ROS to ensure temporal synchronization; raw gaze data from the Aero headset were streamed from Unity to ROS. Analyses were performed offline using Python.

#### *2.3.2 Statistical analyses*

Statistical analyses were performed using Python. Data normality was tested using Shapiro-Wilk test. Data are reported as mean and standard deviation when normally distributed, and as median and interquartile range otherwise. For two-groups comparisons, when both distributions satisfied the normality assumption, the Student's t-test for independent samples was applied; otherwise, the Mann-Whitney U test was used. For three-groups comparisons, if all distributions were normally distributed, a one-way analysis of variance (ANOVA) was performed. When the normality assumption was violated for at least one group, the Kruskal-Wallis test was used as a non-parametric alternative. Post-hoc pairwise comparisons were performed for comparisons involving more than two groups and when the main test was statistically significant ($p < 0.05$): Tukey's honestly significant difference (HSD) test following ANOVA, or Dunn's test with Bonferroni correction following Kruskal-Wallis analysis. To evaluate the relationship between fixation and grasp site positions, correlation analysis was performed. When both variables satisfied the assumptions of normality and linearity, Pearson's correlation coefficient (r) was computed; otherwise, Spearman's rank correlation coefficient ($\rho$) was used as a non-parametric alternative. All statistical tests were two-tailed, with a significance level set at $\alpha=0.05$.

## 3 RESULTS

### 3.1 Gaze-hand coordination in a single target contact trial

Figure 3 shows the pattern of eye and hand movements during a representative trial involving the triangular obstacle, over a time window extending from the reach phase to the end of the exit phase. Gaze and hand movements were temporally coordinated, with gaze leading the hand to important task locations (landmarks). Throughout reach, the participant's gaze was already directed toward the manipulation bar (1) (Figure 3). When the robot's index fingertip contacted the bar, the gaze briefly returned to the hand (2, 3) before shifting to the point where the bar tip was expected to move next (4), and subsequently to the protruding point of the obstacle (5). As the hand lifted the bar, gaze alternated between the predicted trajectory of the bar and the obstacle (6, 7, 8). Once the bar and hand rounded the obstacle during the upward movement, the gaze shifted to the target (9). After the bar had contacted the target, the gaze anticipated the bar's trajectory (11, 12). Finally, as the bar rounded the obstacle again on the downward path, the gaze shifted to the support surface where the bar would be placed (13). In contrast, before the hand enters the camera's field of view and immediately after it leaves the scene (i.e., before the reach and after the exit phase), fixations were spread across the scene without a clear spatial preference for any landmark (Fig. S1).

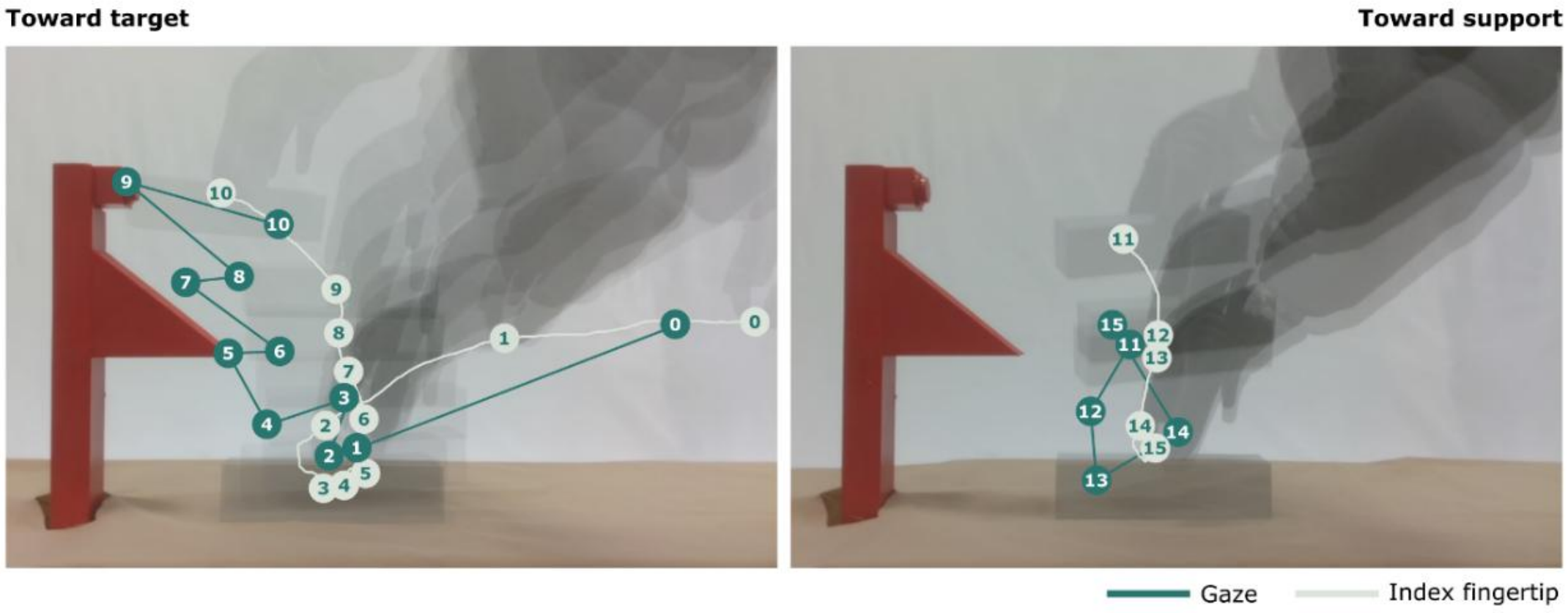

**Fig. 3. Gaze-hand coordination pattern in a representative trial.** Left panel shows movements from *reach* to the end of *to target*, while right panel shows movements from target contact to the end of *exit*. The green line connects fixations, whereas the white line represents the trajectory of the index fingertip. Only fixations whose Euclidean distance from the preceding fixation exceeded the predefined threshold of 24 px were included in the analysis. Green numbered circles indicate the sequence of fixations, and white numbered circles represent the sequence of index fingertip positions at corresponding times. The background image was generated by extracting and overlaying the first video frames corresponding to each selected fixation.

### 3.2 Fixation landmarks

Participants directed their gaze almost exclusively to task-relevant landmarks, including the grasp site, left tip of the bar, protruding point(s) on the obstacle, the target, support surface, and the hand. Figure 4A shows fixations relative to landmarks from all participants and trials, over a time window from the pre-entry phase to the end of the reset phase. Across conditions, fixations were scattered within the grasp site and the tip of the bar. At the reach onset, gaze was directed to the hand, whereas fewer fixations on the hand were observed at the end of exit. In NONE, gaze was directed to the target both before and after the bar tip entered this area. In contrast, in RECT and TRI, gaze shifted toward the target mainly after the bar had rounded the obstacle. A total of 4395 fixations were recorded, corresponding to an average of approximately 33 fixations per trial. Most fixations occurred on landmarks (3313 of 4395, 75%), increasing to 87% when considering the reach-to-exit time window. Fixation duration within landmarks (median, 0.28 s; IQR, 0.44 s) did not differ significantly from that outside landmarks (median, 0.29 s; IQR, 0.34 s) (Mann-Whitney U test; $p = 0.135$). However, a different pattern emerged when focusing on the reach-to-exit time window, with fixations on landmarks (median, 0.28 s; IQR, 0.49 s) significantly longer than those outside landmarks (median, 0.22 s; IQR, 0.25 s) (Mann-Whitney U test; $p < 0.001$). On average, participants directed their gaze toward landmarks for 51% of the total time spent looking at the scene, increasing to 57% when considering the time window from reach to the end of exit, consistent with previous results obtained when considering the same time window.

Fixations at the grasp site, tip of the bar, obstacle and target were not uniformly distributed within their respective landmarks. Instead, they formed clusters whose degree of clustering differed across landmarks (Figure 4B). The target showed the smallest cluster diameter (7.64° - 6.68 cm), compared with the grasp site (10.82° - 9.48 cm), the tip of the bar (9.24° - 8.08 cm), and the obstacle, whose cluster diameter was similar whether computed across the entire trial (9.02° - 7.88 cm) or computed by specifically considering the period before (8.9° - 7.78 cm) and after (9.12° - 7.98 cm) the tip of the bar approaches the target. The centers of fixation clusters did not coincide with the nominal landmark locations; all distances between clusters centers and landmarks centers were computed as Euclidean distances. For the target, the cluster center was located 9.8 mm from the center of the target's contact surface, and for the obstacle it was also 9.8 mm from the obstacle tip. When considering fixations directed to the obstacle before and after the bar tip entered the target, the cluster centers were 10.1 mm and 9.8 mm from the obstacle tip, respectively. For the grasp site, the center of fixations was located 15.6 mm from the grasp location, while for the bar tip, it was 8 mm away.

Figure 4C shows the grasp site location in bar coordinates and the corresponding fixations positions during the interval from the pre-entry phase to the start of the adjustment phase. Participants fixated the grasp site in 25% of the trials. In these trials, fixations and grasp site positions were positively correlated ($r = 0.73$; $p < 0.001$), with a slope close to 1, indicating that fixations predicted the forthcoming grasp position. The percentage of trials with fixations on the grasp site increased to 53% when the area of interest included the entire bar.

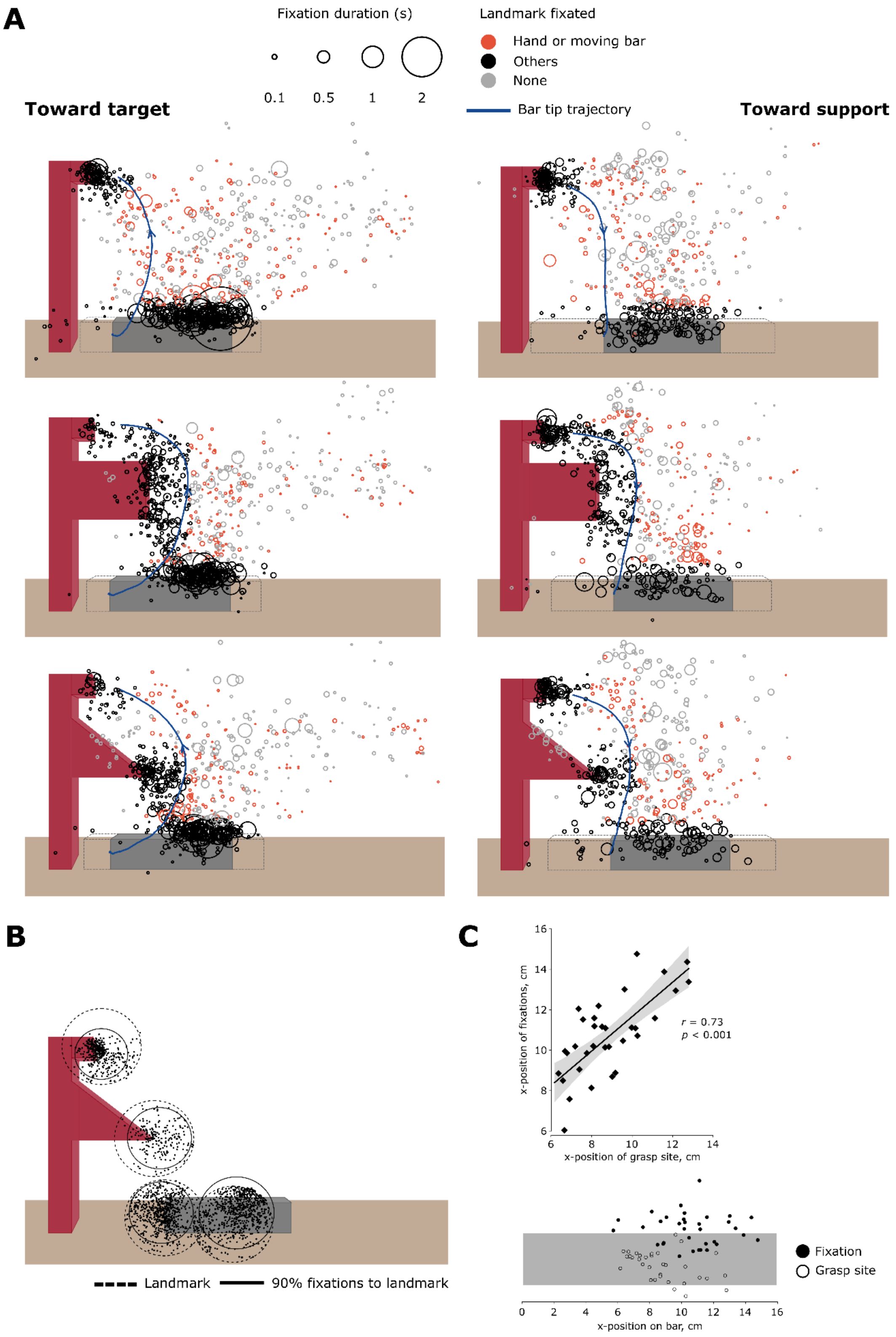
A
Fixation duration (s)
0.1
0.5
1
2
Landmark fixated
Hand or moving bar
Others
None
Bar tip trajectory
Toward target
Toward support
B
Landmark
90% fixations to landmark
C
r = 0.73
p < 0.001
x-position of fixations, cm
x-position of grasp site, cm
Fixation
Grasp site
x-position on bar, cm

**Fig. 4. Fixations relative to landmarks across conditions.** (A) Distribution of fixations from all participants and trials, shown separately for each obstacle condition. Left panels show fixations from *pre-entry* to the end of *up*; right panels show fixations from *to target* to the end of *reset*. Fixations within stationary landmarks are shown in black, within moving landmarks in orange, and outside landmarks in grey. Circle area scales with fixation duration. The bar is shown at its mean position at the beginning and end of the trial, with dashed extensions indicating inter-trial variability. Dispersion of fixations within the grasp site and at the tip of the bar when the bar rested on the table reflects trial-by-trial variability in bar horizontal position. The green curve shows the mean trajectory of the bar tip. (B) Distribution of fixations (black dots) within landmark zones. Fixations from all obstacle conditions and trial phases were combined, except for fixations in the obstacle which were taken from the TRI condition. Fixations directed to the grasp site and the tip of the bar were normalized to account for trial-by-trial variability in grasp location and bar position. Specifically, grasp site fixations were referenced to the mean grasp site for all trials by all participants. Dashed lines denote landmarks, and solid lines represent a circle centered on the mean fixation location (x, y), whose diameter is defined to capture 90% of fixations within each zone. (C) Grasp site fixations between *pre-entry* and the start of *adjustment* are shown for trials in which the grasp site was fixated at least once (25% of trials). White dots represent grasp site positions for individual trials, and black dots the corresponding mean fixation positions, expressed in bar coordinates. The scatterplot shows grasp site position as a function of fixation positions.

### 3.3 Sequence of landmarks fixated

Figure 5 shows how participants shifted their gaze between landmarks across all conditions, considering trials by all participants and the time window from the pre-entry phase to the end of the reset phase. In NONE, the target was predominantly the first landmark fixated, whereas in the obstacle conditions the obstacle itself was most often fixated first. Across all conditions, no dominant paths were observed; instead, frequent transitions, including many bidirectional shifts, occurred. The median number of gaze shifts between landmarks per trial was 13 (IQR, 10) in NONE, 15 (IQR, 7) in TRI and 17 (IQR, 8) in RECT.

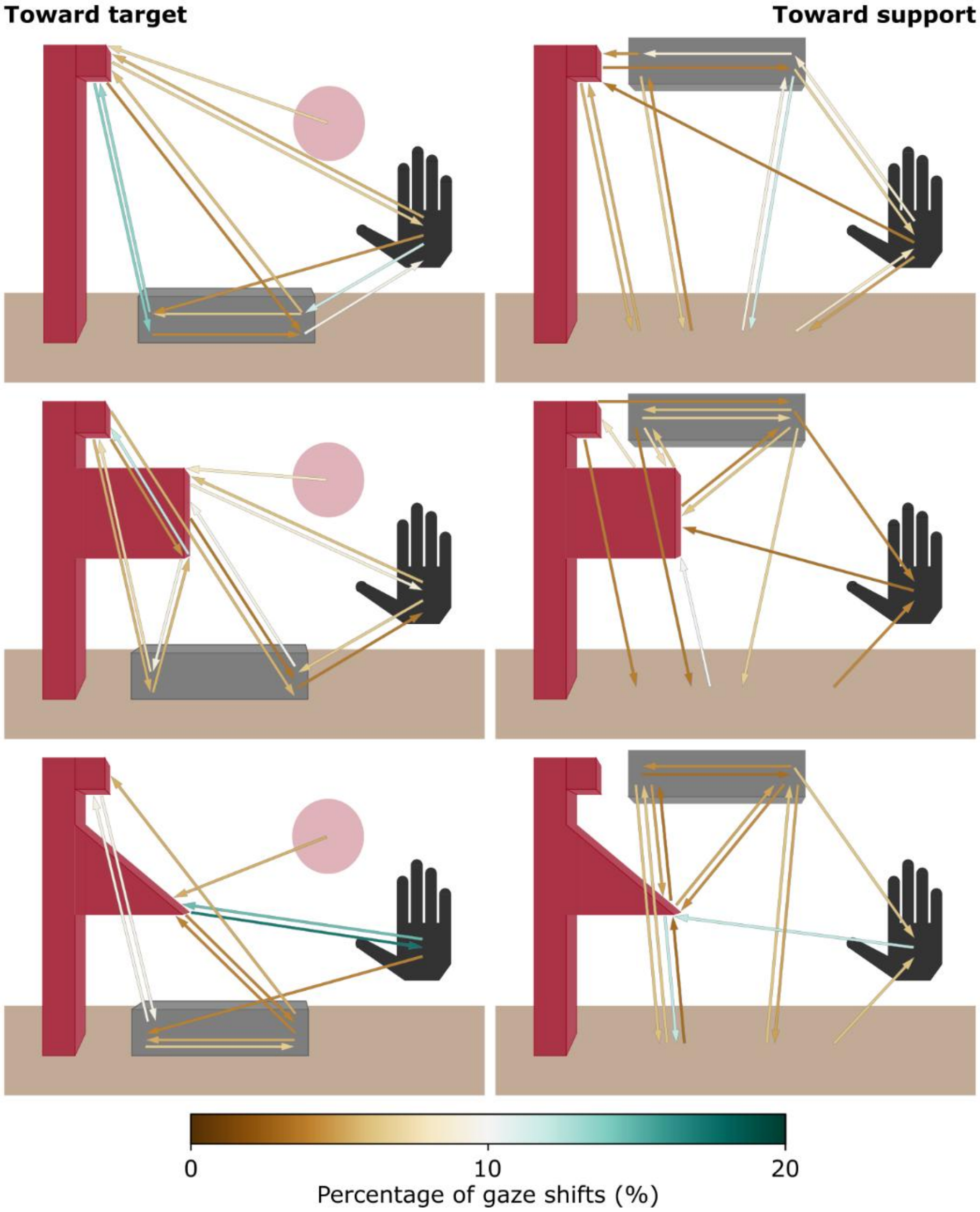

**Fig. 5. Sequence of landmarks fixated across conditions, considering trials by all participants.** A gaze shift was defined as the transition between two consecutive landmark-related fixations. When a non-landmark fixation occurred between them, the gaze shift was considered between the preceding and the following landmark-related fixations. Arrows color indicates the proportion of fixation shifts between the two connected landmarks; shifts at proportions less than 3% are not shown. Left panels show gaze shifts from pre-entry until the end of to target. The red circle was centered on the mean starting gaze coordinates computed across trials, and its radius was defined as the 90th percentile of the Euclidean distances of individual fixations from that mean position. The right panels show gaze shifts from from target until the end of reset. The support surface landmark was considered only for downward movements.

### 3.4 Spatiotemporal coordination of gaze and hand

Figure 6 shows the spatiotemporal coordination of gaze and manipulatory actions using data from all participants and trials involving the triangular obstacle, considering the time window from the pre-entry phase to the end of the reset phase. Figure 6B shows scattered fixations during pre-entry and entry with no landmarks consistently attracting the gaze. By the end of pre-entry, few trials showed fixations on task-relevant landmarks: the bar tip and obstacle in approximately 5% and 32% of trials, respectively; the target in about 18%; and the grasp site and support in 16% and 36%, respectively. During reach, gaze was mainly directed toward the hand, and it gradually shifted to the grasp site during adjustment (Figure 6A, Figure 6B, a, f). This shift continued into grasp, when gaze alternated between grasp site and hand (Figure 6A, Figure 6B, a, f). After the bar moved, gaze never shifted directly to the target (Figure 6A, d). Instead, it first moved to the obstacle and later to the target, typically around the middle of up. Accordingly, the probability of fixating the obstacle increased as that of fixating the grasp site decreased (Figure 6A, a, c), while the probability of fixating the target increased as the probability of fixating the obstacle decreased (Figure 6A, c, d). In contrast, in NONE, the probability of fixating the bar tip and the target increased as that of fixating the grasp site decreased (Figure S2A). Gaze entered the target approximately 1.8 s before the bar tip contacted the target and left about 0.16 s before it moved away (Figure 6B, d). A comparable anticipatory behavior was observed in RECT (Figure S3B). In NONE, anticipation was stronger: gaze entered about 2.4 s before contact and left about 0.15 s before movement away (Figure S2B). Participants never tracked the tip of the bar while resting on the support, but they did track it during bar movement (Figure 6A, b). Specifically, the probability of fixating the bar tip increased as the tip approached the target (Figure 6A, b). Additionally, during the target phase the probability of fixating the grasp site also increased (Figure 6A, a). After leaving the target, gaze never shifted directly from the target to the support, but it was instead directed to all landmarks (Figure 6A). This pattern is reflected in the increased probability of fixating all landmarks during down. Particularly, the probability of fixating the support when replacing the bar was approximately 0.6 (Figure 6A, e), with comparable values for grasp site and hand, and slightly lower for the bar tip (Figure 6A, a, b, f). Because support and grasp site landmarks overlapped, support fixation probability was relatively high before the bar movement. However, only about 10% of fixations were directed to the support landmark alone. Similar to the beginning of the trial, fixations during the reset phase were scattered, with no particular landmark attracting the gaze.

Fixation duration varied across phases (Figure 6C), with longer fixations when gaze was directed toward the grasp site, support surface, target, or hand. Across the entire trial, participants made a median of 26 fixations per trial (Figure 6D).

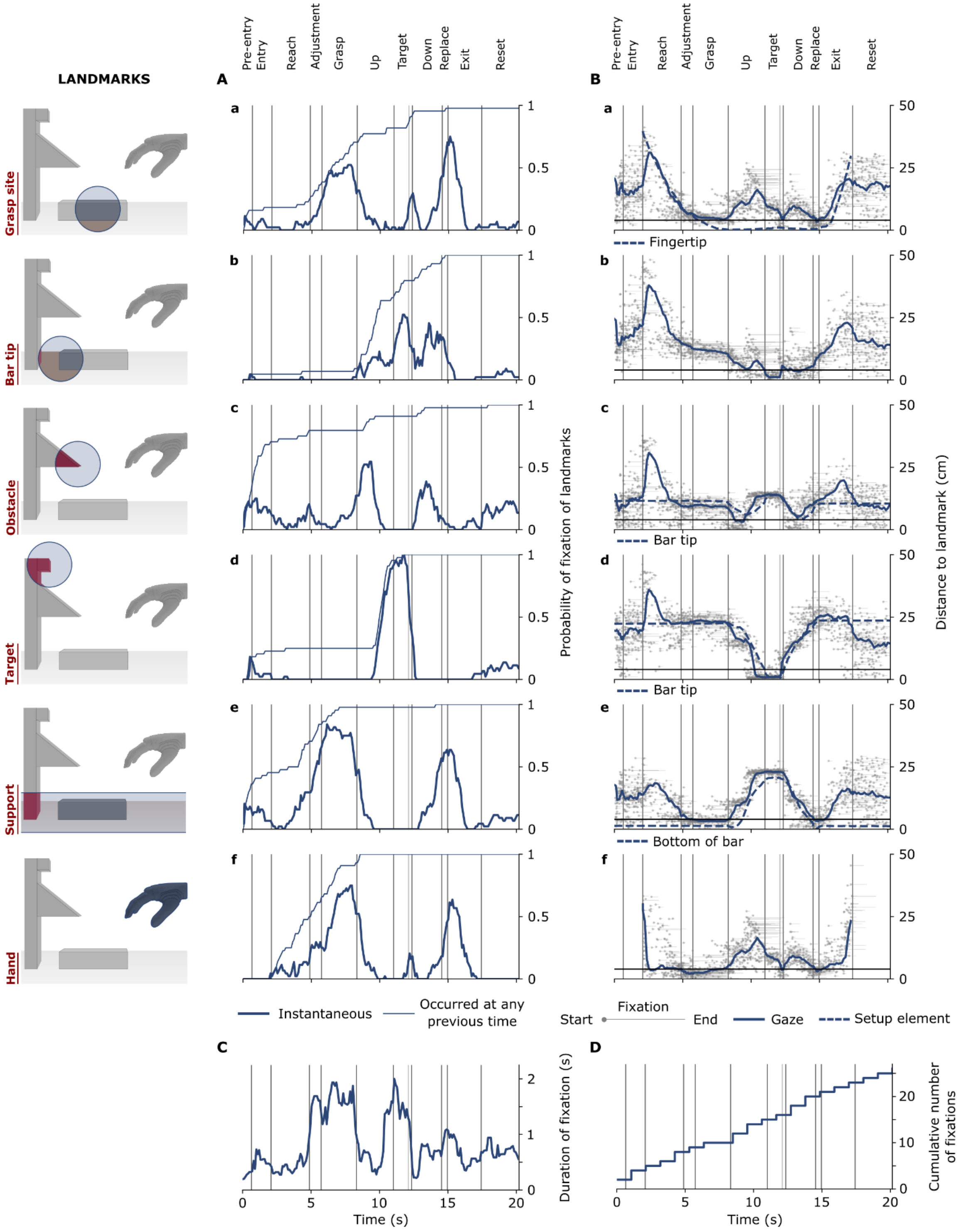
LANDMARKS
A
B
Pre-entry
Entry
Reach
Adjustment
Grasp
Up
Target
Down
Replace
Exit
Reset
Grasp site
Bar tip
Obstacle
Target
Support
Hand
a
b
c
d
e
f
Fingertip
Bar tip
Bar tip
Bottom of bar
Probability of fixation of landmarks
Distance to landmark (cm)
Instantaneous
Occurred at any previous time
Start
Fixation
End
Gaze
Setup element
C
D
Time (s)
Time (s)
Duration of fixation (s)
Cumulative number of fixations

**Fig. 6. Spatiotemporal coordination of gaze and manipulatory actions.** The analysis includes data from all participants and trials involving the triangular obstacle. Vertical lines mark task phase transitions, with phase labels shown above; target phase combines the *to target* and *from target* phases. The time axis was normalized by scaling each phase of each trial to the median phase duration. (A) Fixation probabilities for the landmarks. The thick line shows the instantaneous probability of fixations within the landmark (100 ms bins), and the thin line the probability that a fixation within a landmark had occurred previously during the trial. (B) Spatiotemporal relationship between gaze and setup elements relative to landmarks. The solid line shows the distance between the median gaze position and the landmark. Dots indicate gaze positions at fixation onset, with horizontal lines indicating fixation duration. The dashed line shows the distance between the median position of the index fingertip (a), bar tip (c, d), or bottom of the bar (i.e., lowest corner) (e) and the landmark. Horizontal rectangles indicate the 4 cm landmark zones. (C) Median duration of fixations and (D) cumulative median number of fixations per trial over normalized time.

### 3.5 Fixation parameters across landmarks and conditions

Figure 7A-D illustrates fixations parameters across landmarks and conditions, from the pre-entry phase to the end of the reset phase.

Figure 7A shows the probability of fixating each landmark during a trial across obstacle conditions. Across all conditions, the grasp site, bar tip, obstacle up, target, and hand were fixated in almost every trial and therefore can be considered obligatory gaze landmarks. In contrast, obstacle down and support surface landmarks were fixated less often and can be considered optional landmarks.

The total duration of fixations at the grasp site was reduced in the presence of an obstacle. Specifically, it was significantly longer in NONE than in RECT (Kruskal-Wallis test; $p < 0.05$) (Figure 7B). However, obstacle presence did not significantly affect the number of fixations directed to the grasp site. The target landmark showed the opposite pattern, with significantly more fixations in NONE than in both obstacle conditions (Kruskal-Wallis test; RECT: $p < 0.01$, TRI: $p < 0.001$) (Figure 7C). During the up phase, both total duration and number of fixations on the obstacle were significantly greater in RECT than in TRI (Mann-Whitney U test; duration: $p < 0.05$; number: $p < 0.01$) (Figure 7B, Figure 7C). Fixations on the support surface also lasted significantly longer in NONE than in RECT (Kruskal-Wallis test; $p < 0.05$) (Figure 7B).

Figure 7D shows that individual fixation duration at the bar tip was significantly longer in NONE than in RECT (Kruskal-Wallis test; $p < 0.05$). Similarly, individual fixations on the hand lasted significantly longer in NONE than in RECT (Kruskal-Wallis test; $p < 0.05$).

Several movement phases also differed in their duration across conditions (Figure 7E). The pre-reach phase was significantly longer in NONE than in RECT (Kruskal-Wallis test; $p < 0.05$). In the up phase a significant main effect of obstacle condition was found (Kruskal-Wallis test; $p < 0.05$) although post-hoc pairwise comparisons were not significant. Lastly, the exit phase was significantly longer in NONE than in RECT (ANOVA; $p < 0.01$).

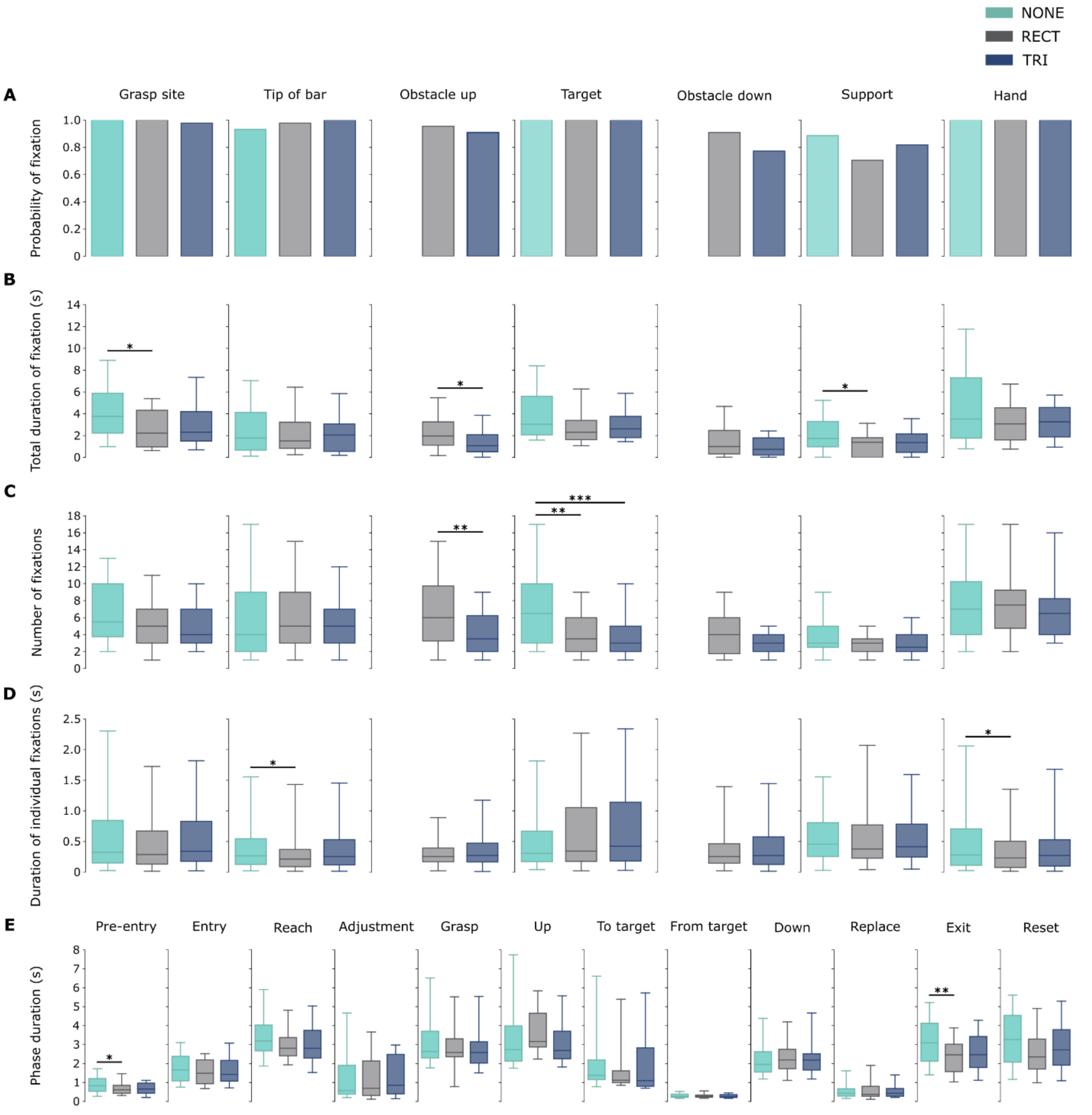

**Fig. 7. Characteristics of gaze fixations and durations of movement phases across conditions.** The analysis combines data across all participants and trials. Mint green, grey and blue bars refer to the NONE, RECT and TRI conditions, respectively. Fixations on the obstacle before and after the up phase are shown separately. Fixations on the support surface refer to those occurring after the up phase. (A) Probability of fixating the landmark during a trial. (B) Total fixation duration for each landmark. (C) Number of fixations per landmark for trials in which that landmark was fixated. (D) Durations of individual fixations for each landmark. (E) Duration of each phase of the trial. Boxplots show the median and interquartile range (25th - 75th percentiles), with whiskers extending to the 5th and 95th percentiles. Outliers are not shown to improve figure readability. Horizontal bars represent significant differences obtained with post-hoc tests (* for $p < 0.05$, ** for $p < 0.01$, *** for $p < 0.001$).

### 3.6 Time relations between gaze shifts and kinematic events

Figure 8 shows the temporal relationship between kinematic events at each landmark and the moments when gaze entered or exited that landmark, from the pre-entry phase to the end of the reset phase. For each landmark, cumulative frequency of trials were computed from the relative timing of gaze entry and exit times with respect to the kinematic event, including only trials in which the landmark was fixated before the event.

For the grasp site (Figure 8, a), cumulative frequency distributions were computed from 32 trials in NONE, 28 in RECT, and 30 in TRI. Gaze reached the grasp site significantly earlier in NONE than in RECT (Kruskal-Wallis test; $p < 0.05$). Specifically, gaze entered this landmark 5 s before index finger contact with the bar in NONE (SD, 3.1 s), compared with 1.8 s (IQR, 3.6 s) in RECT and 2.7 s (IQR, 5.6 s) in TRI. Regardless of condition, gaze typically exited the grasp site before finger contact (NONE: median, 0.5 s; IQR, 2.5 s; RECT: median, 0.3 s; IQR, 0.6 s; TRI: median, 0.5 s; IQR, 1.4 s), though in some trials it exited after contact. Importantly, obstacle presence did not affect gaze exit timing (Kruskal-Wallis test; $p = 0.866$).

For the hand during reaching (Figure 8, b), cumulative frequency distributions were computed from 41 trials in NONE and 44 in both RECT and TRI. Gaze entered the hand 3.8 s before index finger contact with the bar in NONE (SD, 4 s), compared with 4 s (IQR, 3 s) in RECT and 3.5 s (IQR, 4.5 s) in TRI. In most trials, gaze remained on the hand beyond finger contact (NONE: median, 0.6 s; IQR, 1 s; RECT: median, 0.5 s; IQR, 1 s; TRI: median, 0.5 s; IQR, 0.8 s). No statistically significant differences were observed among conditions for either gaze entry or exit times (Kruskal-Wallis test; entry: $p = 0.548$; exit: $p = 0.486$).

For the obstacle during the up phase (Figure 8, c), cumulative frequency distributions were computed from 42 trials in RECT and 40 in TRI. Gaze reached this zone before the bar approached the obstacle, arriving 9.5 s (SD, 4.6 s) and 9.4 s (SD, 4.9 s) before the closest bar-obstacle distance in RECT and TRI, respectively, with large interquartile ranges indicating high temporal variability across trials. In contrast, gaze exit occurred around the time of closest approach and differed across conditions. In TRI, gaze exited the obstacle slightly before the closest approach (median, 0.2 s; IQR, 0.4 s), whereas in RECT it exited after this event (median, 1 s; IQR, 1.1 s). This difference in gaze exit timing between obstacle conditions was statistically significant (Mann–Whitney U test; $p < 0.001$).

For the target (Figure 8, d), cumulative frequency distributions were computed from 44 trials in NONE, 43 in RECT, and 43 in TRI. Gaze preceded button release across conditions, reaching the target a median of 6.2 s before release in NONE (IQR, 13.5 s), and 2.9 s (IQR, 11.5 s) and 3.4 s (IQR, 7.7 s) in RECT and TRI, respectively. However, differences in gaze arrival times across conditions did not reach statistical significance (Kruskal–Wallis test; $p = 0.054$). Here as well, the large interquartile ranges indicate substantial trial-by-trial variability in gaze timing across all conditions. In contrast, gaze exit was tightly coupled to target interaction. Specifically, gaze exited the target 0.2 s (IQR, 0.4 s), 0.05 s (IQR, 0.3 s), and 0.04 s (IQR, 0.2 s) before the bar released the target in NONE, RECT, and TRI, respectively. Differences among conditions were not statistically significant (Kruskal-Wallis test; $p = 0.078$).

For the obstacle during the down phase, entry and exit time distributions were narrowed around the relevant kinematic event (Figure 8, e). Cumulative frequency distributions were computed from 27 trials in RECT and 22 in TRI. Gaze reached the obstacle before the bar tip approached the obstacle, arriving at 1.2 s (SD, 0.7 s) and 0.7 s (IQR, 0.8 s) before the closest bar-obstacle distance in the RECT and TRI, respectively. Importantly, a statistically significant difference was observed between RECT and TRI (Mann–Whitney U test; $p < 0.05$). Gaze exited the obstacle 0.3 s (SD, 1.5 s), and 0.5 s (SD, 0.7 s) after the bar reached the closest bar-obstacle distance in RECT and TRI, respectively.

For the support surface (Figure 8, f), cumulative frequency distributions were computed from 39 trials in NONE, 31 in RECT, and 34 in TRI. Gaze reached the support surface before the bar contacted the surface, arriving at 1.6 s (SD, 1.2 s), 1.2 s (SD, 0.9 s), and 1.2 s (SD, 0.8 s) before the contact in NONE, RECT and TRI, respectively. However, differences in gaze arrival times across conditions were not statistically significant (ANOVA; $p = 0.222$). Gaze exited the support surface after the bar contacted the surface, at 1 s (IQR, 1.1 s), 0.7 s (SD, 0.6 s), and 0.7 s (IQR, 0.8 s) in NONE, RECT and TRI, respectively. Again, no significant differences were observed among conditions (ANOVA; $p = 0.422$).

For the hand (Figure 8, g), cumulative frequency distributions during task completion were computed from 42 trials in NONE, 37 in RECT, and 40 in TRI. Gaze entered this landmark 0.5 s before the bar contacted the surface in NONE (IQR, 1.8 s), compared with 0.2 s (IQR, 0.8 s) in RECT and 0.4 s (IQR, 1.1 s) in TRI. In most trials, gaze remained on the hand beyond the bar contacted the surface (NONE: median, 1.2 s; IQR, 1.4 s; RECT: median, 0.7 s; IQR, 1 s; TRI: mean, 1.1 s; SD, 0.8 s). No statistically significant differences among conditions were observed for either gaze entry or exit times (Kruskal-Wallis test; entry: $p = 0.145$; exit: $p = 0.091$).

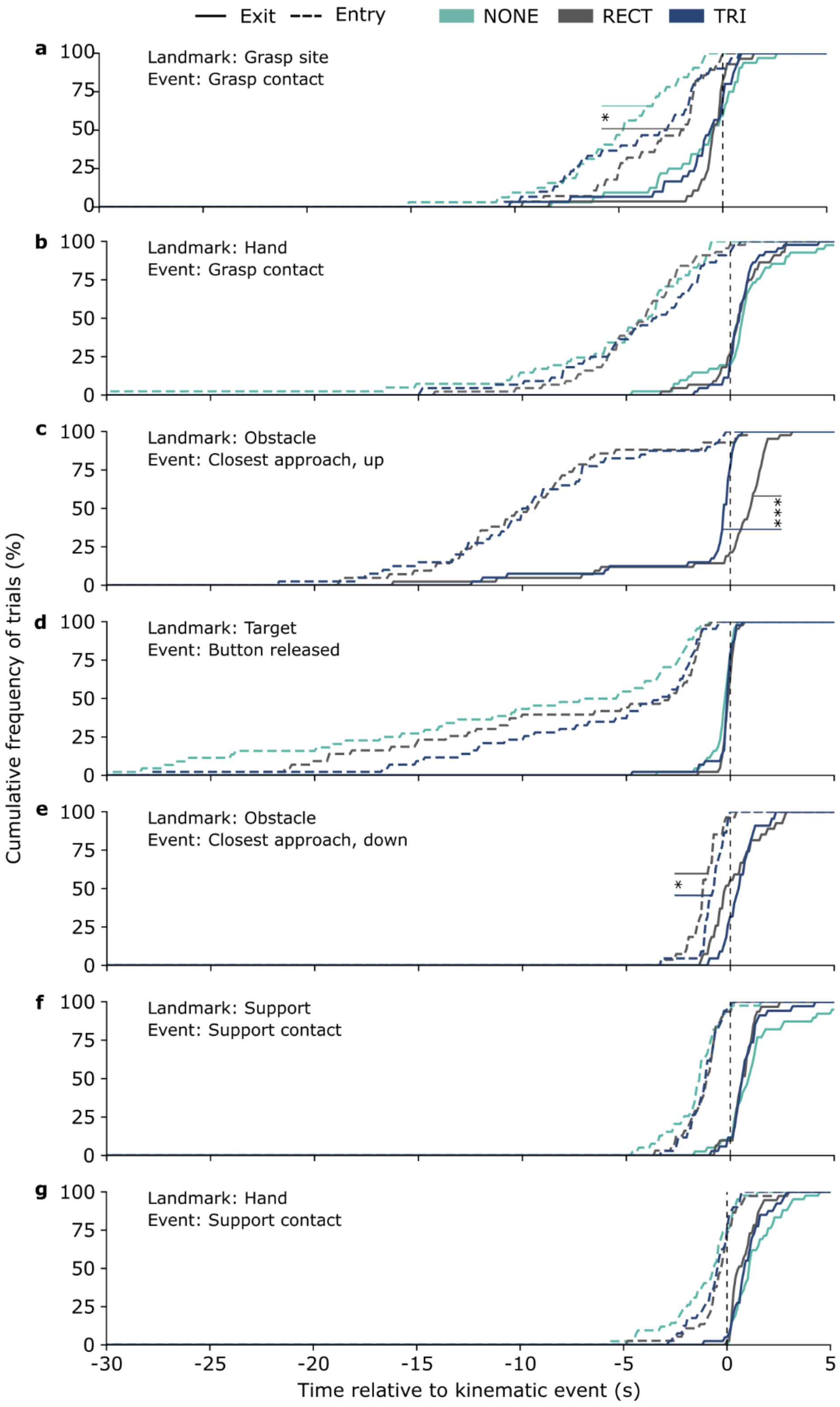

Exit
Entry
NONE
RECT
TRI
a
Landmark: Grasp site
Event: Grasp contact
b
Landmark: Hand
Event: Grasp contact
c
Landmark: Obstacle
Event: Closest approach, up
d
Landmark: Target
Event: Button released
e
Landmark: Obstacle
Event: Closest approach, down
f
Landmark: Support
Event: Support contact
g
Landmark: Hand
Event: Support contact
Cumulative frequency of trials (%)
Time relative to kinematic event (s)

**Fig. 8. Time relations between gaze shifts and kinematic events.** Temporal relationship between kinematic events associated with each landmark and the moments when gaze entered or exited the landmark, represented as a cumulative frequency distribution using 100 ms bins. Dashed lines represent gaze entering the landmark; solid lines represent gaze exiting the landmark. (a) Gaze entering and exiting the grasp site relative to index finger contact with the bar, from *pre-entry* to the end of *grasp*. (b) Gaze entering and exiting the hand relative to index finger contact with the bar, from *reach* to the end of *up*. (c) Gaze entering and exiting the obstacle relative to the time at which the bar tip passed closest to the obstacle, from *pre-entry* to the end of *up*. (d) Gaze entering and exiting the target relative to button release, from *pre-entry* to the end of *from target*. (e) Gaze entering and exiting the obstacle relative to the time at which the bar tip passed closest to the obstacle, from *down* to the end of *replace*. (f) Gaze entering and exiting the support relative to bar contact with the support surface, from *down* to the end of *exit*. (g) Gaze entering and exiting the hand relative to bar contact with the support surface, from *down* to the end of *exit*. Significance is shown by asterisks (* for $p < 0.05$, ** for $p < 0.01$, *** for $p < 0.001$).

# 4 DISCUSSION

## 4.1 General discussion

In this study we examined how gaze behavior changes when object manipulation is mediated by a robot rather than performed directly with the biological hand. To support a systematic comparison with eye-hand coordination during natural object manipulation, we adapted the experimental protocol from Johansson et al. [4] to a teleoperated setting. We found that, during task execution, participants directed their gaze mainly toward task-relevant locations, with 75% of fixations falling within landmarks. This indicates that gaze remained organized around functionally relevant elements of the task and continued to support object manipulation in a goal-directed manner (Figure 4A, Figure 7A). This pattern is consistent with previous findings demonstrating that gaze is tightly coupled to motor actions [1, 2, 4], showing similarities between gaze allocation during teleoperation and that observed in natural object manipulation. At the same time, participants spent less time fixating task-relevant landmarks than previously reported [4] (51% versus 90% of the total viewing time), suggesting that, although the task-related organization of gaze was preserved, visual attention was redistributed more broadly across the scene. In particular, gaze was not directed exclusively toward upcoming action goals, but was also recruited to monitor dynamic task-relevant elements, including the manipulated object and the robotic hand. Consistent with this interpretation, teleoperation resulted in greater use of smooth pursuit eye movements, indicating an increased need to visually track ongoing movement during task execution (Figure S4). Together, these findings indicate that gaze remains goal-directed during teleoperation, but its functional role is reorganized to support increased online monitoring.

Importantly, the increased reliance on online monitoring did not abolish the predictive nature of gaze. Instead, our findings suggest that the functional role of gaze changes across task phases, dynamically shifting between online monitoring and prediction according to task demands. This interpretation is consistent with the taxonomy proposed by Land et al. [38], according to which gaze serves multiple functions over the course of goal-directed actions.

Before the robotic hand entered the camera field of view, fixations were spatially spread across the scene, without a clear predominance of specific landmarks (Figure S1, Figure 6B). This pattern reflects an exploratory phase in which participants scanned the manipulation environment through locating fixations directed toward landmarks. These initial fixations allow participants to build a spatial representation of the remote environment, which supports later planning of actions. A similar pattern emerged again at the end of the trial, once the robotic hand left the camera field of view, likely reflecting the end of the active visuomotor control.

Between these two phases, gaze function changed over time. Once the robotic hand became visible, fixations shifted from locating to directing function, targeting future task-relevant locations (Figure 2, Figure 3). Importantly, anticipatory fixations were observed both at the target and the grasp site, indicating that teleoperation preserved the predictive component of gaze.

Target-directed anticipatory fixations varied across experimental conditions. In the absence of obstacles, the target became immediately relevant to plan the action, and participants often fixated it early, at the beginning of the upward movement phase (Figure S2A, c). When an obstacle was present, gaze priorities changed: participants first directed gaze to the obstacle, reflecting the urgent need for collision avoidance. As a result, fixations toward the target were delayed and occurred approximately at the middle of the upward phase (Figure 6A, d, Figure S3A, d). Obstacle geometry further modulated gaze temporal organization. Because of its larger size, the rectangular obstacle required higher motor precision than the triangular obstacle, resulting in further delays in gaze shifts toward the target (Figure 6A, d, Figure S3A, d). The greater motor control difficulty associated with the rectangular obstacle was also evident in longer duration and increased number of fixations during the upward phase (Figure 7B, Figure 7C). Overall, the more complex the task, the later participants directed their gaze to the target (Figure 7E). Thus, in obstacle conditions, and in contrast to natural manipulation [4], trajectory planning toward the target was constrained by obstacle avoidance, resulting in anticipatory fixations that alternated between the obstacle and the future position of the manipulation bar (Figure 3).

Anticipatory fixations were also observed at the grasp site (Figure 4), but their temporal dynamics differed from those reported in natural manipulation of the same task [4]. Whereas in natural manipulation gaze typically leaves the object before contact, during teleoperation gaze remained on the manipulation bar throughout grasping. In particular, fixations alternated between the manipulation bar and the robotic hand, likely reflecting the need for visual monitoring to stabilize the grasp and regulate contact forces (Figure 6A, a, f) and suggesting a progressive shift from directing to checking gaze behavior. Consistent with this monitoring role under increased control demands, fixations also alternated between the bar tip and the target during button pressing (Figure 6). Notably, despite the presence of auditory feedback signaling correct button activation, these checking fixations persisted also after button pressing, reinforcing the role of vision as the dominant sensory modality for action monitoring during teleoperation (Figure 4A, right panel). Figure 5 clearly illustrates the checking role of gaze under high control demand, with bidirectional arrows highlighting the absence of a dominant gaze path and, consequently, frequent gaze shifts used for online monitoring of task progress. Additionally, fixations on moving objects were observed throughout the entire trial, consistent with sustained monitoring of action progress. Importantly, natural object manipulation is characterized by predominantly anticipatory gaze [4], with no evidence of checking behavior.

This reorganization is likely related to the altered sensory conditions imposed by teleoperation. Natural object manipulation relies on the cognitive ability to anticipate future states of actions through the integration of multiple sensory modalities, with somatosensory feedback playing a central role in supporting predictive control [39, 34]. When this feedback is degraded or absent, the sensorimotor mechanisms that enable dexterous manipulation are impaired [40]. Under these conditions, humans adapt by reorganizing their sensorimotor strategies [41]. By contrast, teleoperation introduces increased uncertainty about action outcomes, likely because of reduced somatosensory feedback, leading to the emergence of checking fixations. As a result, gaze is not merely anticipatory but dynamically reconfigured across task phases, alternating among exploratory, predictive, and online monitoring functions.

The observed gaze patterns are consistent with prior findings on mediated interactions, where users allocate visual attention not only to task goals but also to the controlled proxy. When manual actions, visual feedback and the controlled object are not spatially co-located, users must rely more heavily on vision to monitor the consequences of their actions. Accordingly, indirect cursor control has been associated with more frequent visual checking of the controlled proxy, whereas direct interactions allow greater reliance on proprioception and more sustained gaze on the task target [10]. Evidence from virtual reality indicates that the absence of haptic feedback can similarly increase visual attention toward the virtual hand and manipulated object [44]. Recurrent fixations on the robotic hand and manipulated object in our study can therefore be interpreted as part of the visual demands of controlling a remote external effector. In particular, the initial fixations on the robotic hand may reflect the need to verify the mapping between motor commands and their visual consequences before predictive gaze is redirected toward future task goals. A similar pattern has also been reported in upper-limb prosthesis use, where disrupted tactile and proprioceptive feedback can increase reliance on vision [41]. Similarly, visuomotor behavior may also vary according to the characteristics of the myoelectric controller [43]. Prosthesis users often alternate gaze between the target object and the prosthetic hand to monitor action execution, and greater visual attention to the hand has been associated with lower skill or confidence in device control [45, 46]. However, anticipatory gaze toward the target is not necessarily lost, as it was found that target fixations still preceded grasp-related actions by approximately 500 ms, even in amputated participants [47]. Together, these findings support the interpretation that reduced sensory feedback can increase online monitoring without eliminating the predictive role of gaze.

Robot-mediated manipulation combines both sources of visual demand: control is exerted through a spatially separate external proxy, and somatosensory information about the remote effector is reduced or absent. We show that, under these conditions, gaze dynamically alternates between proxy monitoring and predictive orienting toward future task-relevant locations, with the relative contribution of each function varying systematically across task phases. These findings have important implications for human-robot interaction by refining the interpretation of gaze during robot-mediated manipulation. Rather than providing a direct readout of the user's intended goal, gaze reflects a dynamic interplay between intention, constraint management, and online monitoring. Accounting for the phase-dependent functional role of gaze is particularly relevant to improving gaze-based intent inference and enabling shared autonomy systems to provide assistance that remains aligned with the user's evolving intentions.

### 4.2 Limitations and future works

Some limitations should be considered when interpreting these results and in view of possible applications. First, the number of task repetitions was deliberately limited to align with the established natural manipulation protocol proposed by Johansson et al. [4], thereby facilitating a structured comparison with prior work. This design choice, however, may not have been sufficient for participants to fully consolidate a stable manipulation strategy. Consequently, the frequency of checking fixations may therefore partly reflect early-stage adaptation to the teleoperation interface. This is consistent with the findings from indirect interaction, where users visually monitor a controlled cursor more frequently when the control mapping is spatially separated from the task target [10], and with behaviors emerging in prosthesis use where increased attention to the artificial effector has been associated with limited sensory feedback, lower confidence, and reduced control proficiency [45, 47]. These consistencies suggest that monitoring fixations may decrease with increasing experience, although the overall task-related organization of gaze is preserved. Second, variability in gaze paths and fixation timing likely reflects both inter-participant and intra-participant differences, highlighting the need for prediction models that account for non-stationary gaze strategies. Third, our study focused on direct teleoperation with a specific embodiment and set-up. Changes in robot autonomy are expected to alter the distribution and functional role of gaze, potentially changing the balance between directing and checking fixations [25, 29, 30]. Finally, restoring somatosensory information could reduce visual reliance and promote more natural sensorimotor control strategies [41, 42]. More generally, gaze behavior is likely shaped not only by feedback availability, but also by the balance and reliability of multisensory information [48, 49]. Similarly, changes in visual embodiment, such as replacing an anthropomorphic hand with a non-anthropomorphic gripper, may alter motor control strategies and gaze allocation [50, 51]. Future work should therefore investigate how these factors interact to recover a gaze that more closely resembles that observed during direct manipulation.

## 5 CONCLUSION

In this study, by adapting the classic natural-manipulation paradigm of Johansson et al. [4] to a robot-mediated context, we provided a structured comparison between natural and teleoperated manipulation, thus systematically characterizing fixation allocation, gaze–hand temporal coordination, and gaze timing relative to task-relevant kinematic events. Our results highlight that the predictive role of gaze is preserved even when actions are executed through a robotic system. Most fixations were directed toward task-relevant locations, and gaze anticipated upcoming action goals. However, teleoperation induces a functional reorganization of gaze behavior, including frequent monitoring fixations toward the robotic hand and manipulated object. This monitoring behavior likely compensates for reduced somatosensory feedback and increased control uncertainty. In addition, task complexity and environmental constraints systematically modulate anticipatory gaze timing, with anticipatory fixations delayed under increased motor demands. In conclusion, this work advances our understanding of visuomotor behavior during teleoperation by showing that gaze is not merely preserved under robot mediation but functionally reconfigured in response to its altered sensorimotor demands. By providing the first direct comparison between teleoperation and natural manipulation, we characterize these phase-dependent dynamics and offer new insights to inform the design of more adaptive human-robot interactions.

## ACKNOWLEDGMENTS

We thank I. Mannari for the support provided in the graphic design of the figures. This work is supported by the Italian Ministry of Research, under the PNNR action "THE-Tuscany Health Ecosystem" – Spoke 9 (project number ECS 00000017).

# A APPENDICES

## A.1 Figures

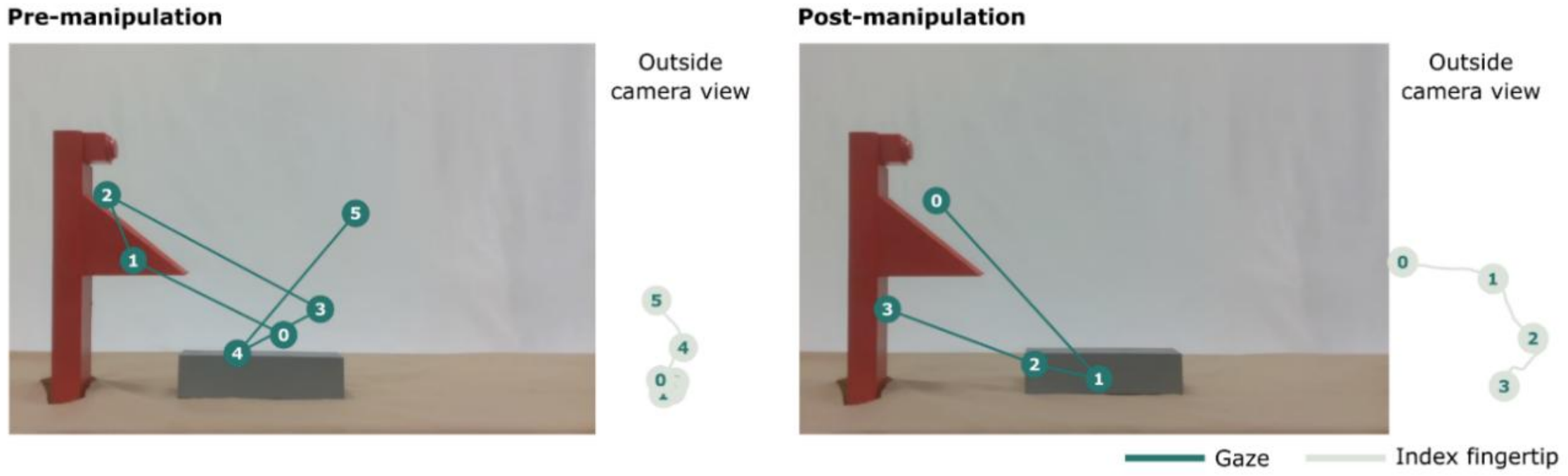

**Fig. S1. Gaze-hand coordination pattern in a representative trial when the hand is not visible to the participant.** Pattern of eye and hand movements during a representative trial with the triangular obstacle when the hand is not visible to the participant. Left panel shows movements from pre-entry to the end of the entry phase, while right panel shows movements during the reset. The green line connects fixations, whereas the white line represents the trajectory of the index fingertip. Only fixations whose Euclidean distance from the preceding fixation exceeded the predefined threshold of 24 px were included in the analysis. Green numbered circles indicate the sequence of fixations, and white numbered circles represent the sequence of index fingertip positions. The background image corresponds to the real scene viewed by the participant through the virtual reality headset.

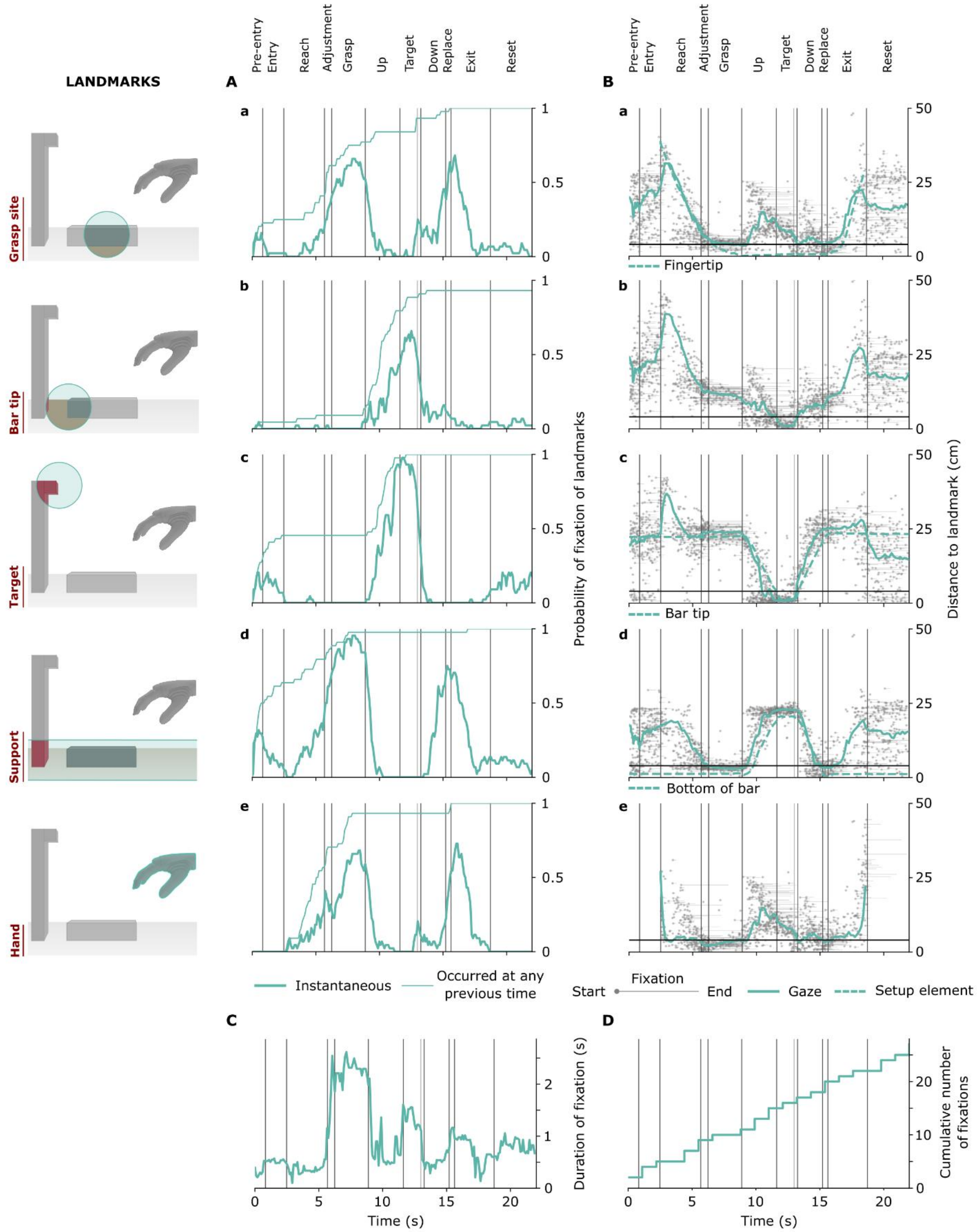
LANDMARKS
A
B
Pre-entry
Entry
Reach
Adjustment
Grasp
Up
Target
Down
Replace
Exit
Reset
Grasp site
Bar tip
Target
Support
Hand
a
b
c
d
e
Probability of fixation of landmarks
Distance to landmark (cm)
Fingertip
Bar tip
Bottom of bar
Instantaneous
Occurred at any previous time
Fixation
Start
End
Gaze
Setup element
C
D
Duration of fixation (s)
Cumulative number of fixations
Time (s)

**Fig. S2. Spatiotemporal coordination of gaze and manipulatory actions.** The analysis includes data from all participants and trials for the NONE condition. Vertical lines mark task phase transitions, with phase labels shown above; target phase combines the to target and from target phases. The time axis was normalized by scaling each phase of each trial to the median phase duration. (A) Fixation probabilities for the landmarks. The thick line shows the instantaneous probability of fixations within the landmark (100 ms bins), and the thin line the probability that a fixation within a landmark had occurred previously during the trial. (B) Spatiotemporal relationship between gaze and setup elements relative to landmarks. The solid line shows the distance between the median gaze position and the landmark. Dots indicate gaze positions at fixation onset, with horizontal lines indicating fixation duration. The dashed line shows the distance between the median position of the index fingertip (a), bar tip (c, d), or bottom of the bar (i.e., lowest corner) (e) and thelandmark. Horizontal rectangles indicate the 4 cm landmark zones. (C) Median duration of fixations and (D) cumulative median number of fixations per trial over normalized time.

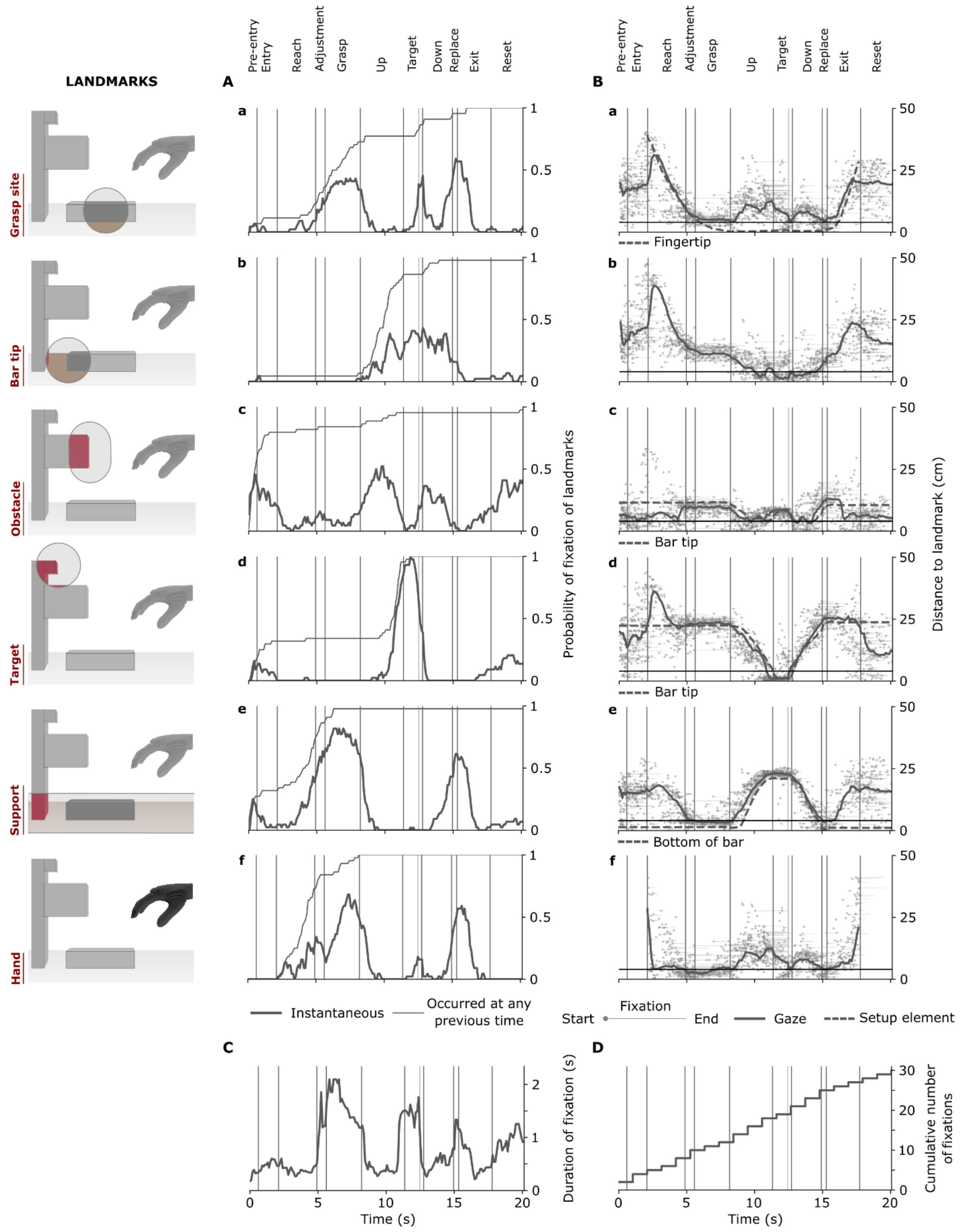

LANDMARKS
A
B
C
D
Pre-entry
Entry
Reach
Adjustment
Grasp
Up
Target
Down
Replace
Exit
Reset
Grasp site
Bar tip
Obstacle
Target
Support
Hand
Fingertip
Bar tip
Bar tip
Bottom of bar
Probability of fixation of landmarks
Distance to landmark (cm)
Instantaneous
Occurred at any previous time
Fixation
Start
End
Gaze
Setup element
Duration of fixation (s)
Cumulative number of fixations
Time (s)
Time (s)

**Fig. S3. Spatiotemporal coordination of gaze and manipulatory actions.** The analysis includes data from all participants and trials for the RECT condition. Vertical lines mark task phase transitions, with phase labels shown above; target phase combines the *to target* and *from target* phases. The time axis was normalized by scaling each phase of each trial to the median phase duration. (A) Fixation probabilities for the landmarks. The thick line shows the instantaneous probability of fixations within the landmark (100 ms bins), and the thin line the probability that a fixation within a landmark had occurred previously during the trial. (B) Spatiotemporal relationship between gaze and setup elements relative to landmarks. The solid line shows the distance between the median gaze position and the landmark. Dots indicate gaze positions at fixation onset, with horizontal lines indicating fixation duration. The dashed line shows the distance between the median position of the index fingertip (a), bar tip (c, d), or bottom of the bar (i.e., lowest corner) (e) and the landmark. Horizontal rectangles indicate the 4 cm landmark zones. (C) Median duration of fixations and (D) cumulative median number of fixations per trial over normalized time.

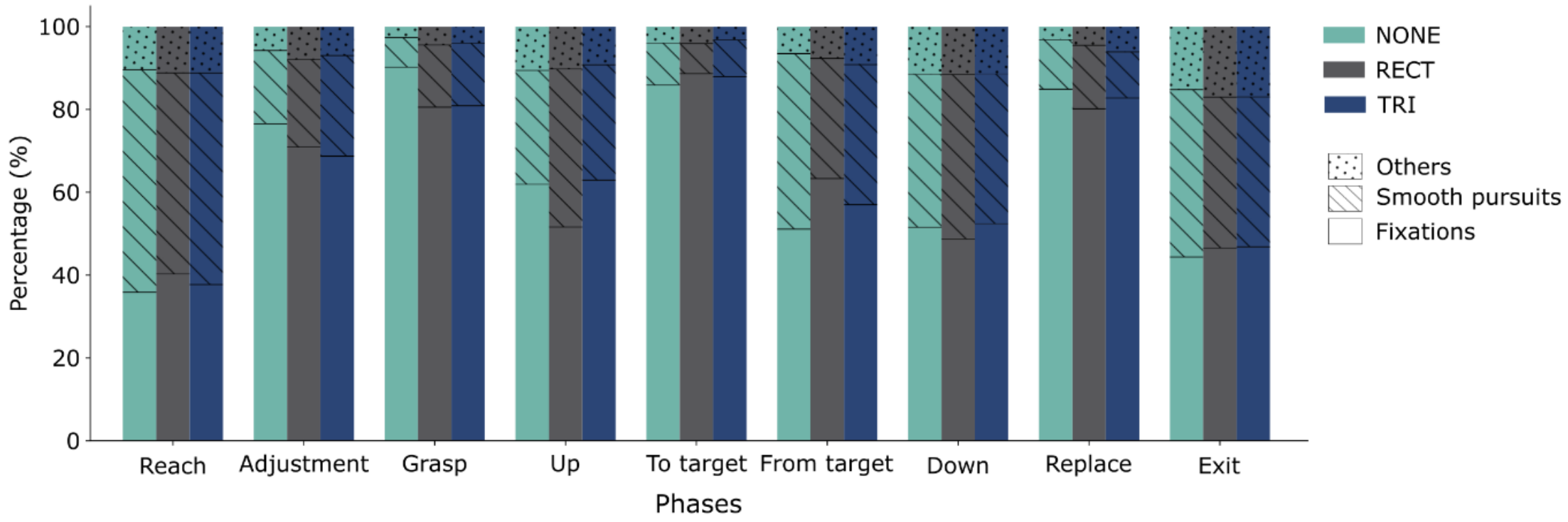

**Fig. S4. Distribution of eye movements across phases and experimental conditions.** The figure shows the percentage of different eye events across phases and experimental conditions. Data are pooled across all participants and trials, from *reach* to the end of *exit*. Textures represent different eye event types (i.e., fixations, smooth pursuit, and others events), while colors indicate the experimental conditions.

## A.2 Movies

Movie S1: Example of a trial execution by a study participant.

Movie S2: Gaze overlay added for visualization on frames shown inside the virtual reality headset during the execution of a trial for each experimental condition.